\documentclass[pmlr,twocolumn,10pt]{jmlr} 

\usepackage{booktabs}
\usepackage{siunitx}

\theorembodyfont{\upshape}
\theoremheaderfont{\scshape}
\theorempostheader{:}
\theoremsep{\newline}

\jmlrvolume{LEAVE UNSET}
\jmlryear{2022}
\jmlrsubmitted{LEAVE UNSET}
\jmlrpublished{LEAVE UNSET}
\jmlrworkshop{Conference on Health, Inference, and Learning (CHIL) 2022} 

\usepackage{multirow}
\usepackage{multicol}
\usepackage{chngpage}
\usepackage{bbold}
\usepackage{amsmath, amssymb}

\title[Lead-agnostic SSL for Local and Global Representations of ECG ]{Lead-agnostic Self-supervised Learning for \\ Local and Global Representations of Electrocardiogram}

\author{%
\Name{Jungwoo Oh} \Email{ojw0123@kaist.ac.kr}\\
\Name{Hyunseung Chung} \Email{hs\_chung@kaist.ac.kr}\\
\addr KAIST, Republic of Korea
\AND
\Name{Joon-myoung Kwon} \Email{cto@medicalai.com}\\
\Name{Dong-gyun Hong} \Email{dghong@medicalai.com}\\
    \addr Medical AI Inc., Republic of Korea
\AND
\Name{Edward Choi} \Email{edwardchoi@kaist.ac.kr}\\
\addr KAIST, Republic of Korea
}

\begin{document}

\maketitle

\begin{abstract}

In recent years, self-supervised learning methods have shown significant improvement for pre-training with unlabeled data and have proven helpful for electrocardiogram signals.
However, most previous pre-training methods for electrocardiogram focused on capturing only global contextual representations.
This inhibits the models from learning fruitful representation of electrocardiogram, which results in poor performance on downstream tasks.
Additionally, they cannot fine-tune the model with an arbitrary set of electrocardiogram leads unless the models were pre-trained on the same set of leads.
In this work, we propose an ECG pre-training method that learns both local and global contextual representations for better generalizability and performance on downstream tasks.
In addition, we propose random lead masking as an ECG-specific augmentation method to make our proposed model robust to an arbitrary set of leads.
Experimental results on two downstream tasks, cardiac arrhythmia classification and patient identification, show that our proposed approach outperforms other state-of-the-art methods.



\end{abstract}
\paragraph*{Data and Code Availability}
This paper uses the \textbf{Physionet 2021} dataset and \textbf{PTB-XL} dataset, which are publicly available on the PhysioNet repository \citep{reyna2021cinc, wagner2020physionet}. More details about datasets can be found at \sectionref{sec:dataset}.

Our implementation code can be accessed at this repository.\footnote{\url{https://github.com/Jwoo5/fairseq-signals}}


\section{Introduction}
\label{intro}
Electrocardiogram (ECG) is the most commonly used measurement to investigate a patient's cardiac activity. It is a non-invasive method to assess the functionality of the heart, which helps to diagnose numerous heart diseases such as arrhythmia, myocardial infarction, and arterial disease.
In any healthcare facility, symptoms related to the heart involve ECG measurement, which leads to the everyday accumulation of ECG recordings.
In previous deep learning methods~\citep{kachuee2018ecg, yan2019fusing}, supervised learning of ECG data was performed with labels only clinical practitioners or professional cardiologists could annotate.
This limited the number of samples available for training and also became a burdensome task for practitioners.

To overcome this limitation, researchers started to adopt the self-supervised learning strategies from other domains such as natural language processing \citep{kenton2019bert, peters1802deep}, computer vision \citep{chen2020simple,grill2020bootstrap}, and speech processing \citep{oord2018representation, baevski2020wav2vec}, where a model is first pre-trained on a large unlabeled dataset, then fine-tuned on a smaller task-specific labeled dataset.
In \cite{kiyasseh2021clocs}, the authors present a self-supervised learning framework that encourages ECG representations across patients to be similar or different from one another.
Similarly, \cite{gopal20213kg} propose generated views using 3D augmentations across positive pairs derived from the same patient to use for contrastive learning.
However, these previous works on self-supervised learning of ECG signals exploit only the global context of ECG recordings, not the local context.
In addition, these methods pre-train and fine-tune under the assumption that ECG data will always consist of the same set of leads for both pre-training and fine-tuning.
However, pre-training and fine-tuning for each lead combination is not a feasible assumption in the real-world environment where it is difficult to know which specific set of leads to be fine-tuned beforehand. 

Therefore, in this work, we propose a contrastive pre-training method considering both global semantic information and the local context of ECG signals. Additionally, we present random lead masking, which enables the model to become robust when only a subset of lead recordings are available at the fine-tuning stage.
Our proposed method outperforms other state-of-the-art ECG pre-training methods for two downstream tasks: cardiac arrhythmia classification and patient identification.
We chose these two tasks because classification and identification require local and global contextualized representations, respectively.
For example, features considering local variability of ECG signals must be extracted to classify cardiac abnormalities, whereas features related to global trends or characteristics of ECG is more important for the identification task.

We summarize our \textbf{contributions} in this work as follows:
\begin{itemize}
\item	We propose an ECG pre-training method that considers both local and global contextual information. To the best of our knowledge, this is the first ECG pre-training approach to consider both together. 
\item	We present Random Lead Masking (RLM), which allows a robust model performance on downstream tasks with an arbitrary number of leads.
\item	In two downstream tasks, diagnosis classification and patient identification, our proposed method consistently achieves stronger performance compared to previous state-of-the-art methods.   
\end{itemize}

\begin{figure*}[t]
\floatconts
    {fig:model}
    {\caption{\textbf{Illustration of our proposed framework.} It jointly learns local and global context by integrating Wav2Vec 2.0 and CMSC. Temporally adjacent segments are passed through the convolutional encoder after some leads are randomly masked. Then, randomly selected latent tokens are replaced with the mask token $m$, as well as quantized to $q$.
    The local contrastive loss is computed between the local representations at masked time steps and their quantized features. The positive pair and negative pairs are shown as blue and red arrows, respectively. To learn global context together, local representations are average-pooled to generate global representations. Then, the network is trained to maximize agreement between temporal adjacent ECG segments.}}
    {\includegraphics[width=1.0\linewidth]{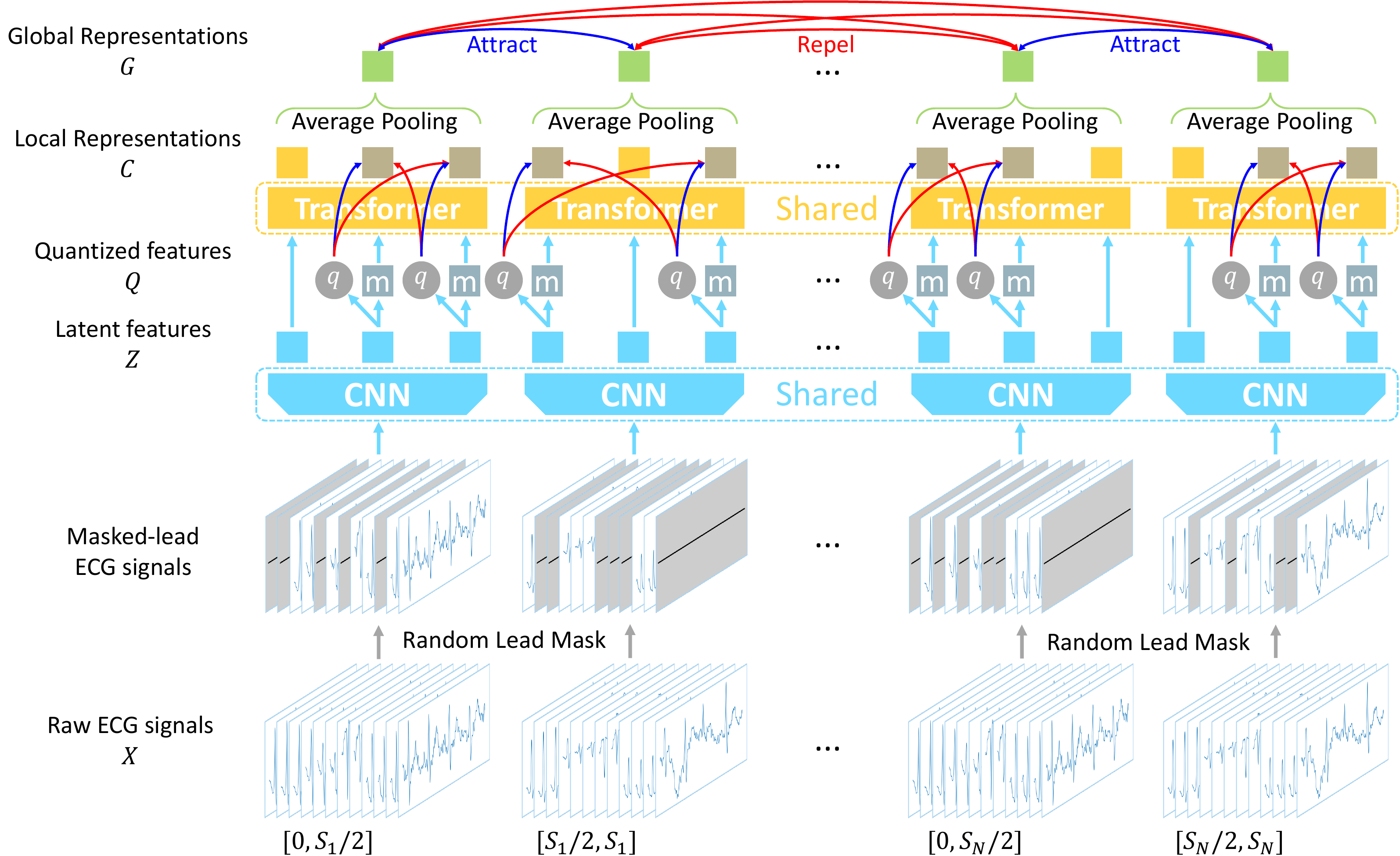}}
\end{figure*}


\section{Related Works}
\label{related}

\paragraph*{Electrocardiogram}
ECG is acquired through several electrodes placed on certain surfaces of the body.
By measuring the voltages between pairs of corresponding attachments, we can evaluate the cardiac function of the heart.
The standard ECG is composed of 12 leads (channels), where each lead measures a specific electrical potential difference.
However, retaining all 12 leads of ECG is not common in the real-world environment because patients must attach at least ten electrodes to the skin in a healthcare facility.
Therefore, many practitioners and cardiac experts utilize reduced-channel systems~\citep{green2007best}.

Previous works show that a deep learning-based approach is effective for automatically diagnosing several heart diseases with ECG.
\citet{kachuee2018ecg} found that deep residual neural networks achieve high accuracy on heartbeat classification tasks and also on MI classification tasks.
Furthermore, \citet{yan2019fusing} developed a heartbeat classifier based on Transformer~\citep{vaswani2017attention}, showing extremely high performance in arrhythmia classification.

Biometric identification with ECG has also been a promising field~\citep{labati2019deep, li2020toward} since common approaches related to personal characteristics, such as fingerprints or facial images, are known to be vulnerable.
For example, facial images can be replicated by artificial masks, and fingerprints can also be easily copied by silicon.
However, since cardiac activity is controlled by the autonomic (involuntary) nervous system, it is difficult to mimic or simulate someone's ECG signal.
Therefore, personal identification using ECG can be an attractive alternative to the previous vulnerable approaches.

\paragraph*{Contrastive self-supervised learning}
Recent works aim to learn high-level global semantic information by designing positive views to be close together and negative views to be far apart in the latent representation space.
In computer vision, \citet{chen2020simple,grill2020bootstrap} propose to define positive pairs as differently augmented views of the same image.
These works demonstrate that data augmentations are highly important for global contrastive learning which is based on perturbations.

In the audio domain, local contrastive learning has been researched in speech processing such that contrastive views are constructed on different frames of the same audio.
For example, \citet{oord2018representation} propose to learn useful representations from auto-regressively predicting the future features in the latent space.
More recently, \citet{baevski2020wav2vec} achieve the best performance in speech recognition through quantization and masking of inputs in the latent space and solving a contrastive task over masked outputs and quantized features.
We utilize this work as the representative model for local contrastive learning since it has been state-of-the-art in the speech domain.

% ...

\paragraph*{Representation learning for electrocardiogram}
Recently, methods for learning representations from ECG via self-supervised learning have been actively developed.
Most of them extend global contrastive learning for ECG.
For example, 3KG \citep{gopal20213kg} introduces several augmentations, especially in the 3-dimensional space of ECG, which is called vectorcardiogram (VCG).
Motivated by the fact that the natural characteristic of a heart varies without critical clinical implications, they convert ECG into VCG and generate positive pairs of VCGs by applying stochastic augmentations such as 3-D rotation and 3-D scaling.
Afterward, they convert the augmented VCGs back to ECGs and solve a global contrastive learning task similar to SimCLR~\citep{chen2020simple}.
\citet{kiyasseh2021clocs} propose another approach to construct effective contrastive views.
They define positive pairs as different spatiotemporal segments of ECGs from the same patient: different frames of the same ECGs (CMSC), different leads of the same ECGs (CMLC), or combining both of them together (CMSMLC).
We use CMSC method as one of the global contrastive methods for ECG since \cite{kiyasseh2021clocs} mentions CMLC performs consistently worse than CMSC.
The author speculates this is due to most heart diseases not being equally visible in all 12 leads at the same time point. On the other hand, CMSC shows superior performance by using non-overlapping temporal segments of the same lead in ECG recordings as positive pairs for contrastive learning.
\section{Pre-training}
In this section, we first describe our contrastive learning strategy.
Specifically, we describe Wav2Vec 2.0 (\textit{i.e.} capturing local features) and CMSC (\textit{i.e.} capturing global features), and the combination of the two.
Then we describe Random Lead Masking (RLM), which helps the model better generalize to downstream tasks with an arbitrary number of ECG leads.

\subsection{Contrastive Learning}
\label{sec:contrastive}
\paragraph*{Wav2Vec 2.0}
Wav2Vec 2.0 consists of several convolutional blocks to encode raw ECG signal inputs and several transformer encoder blocks to derive contextualized local representations from the CNN outputs. 
Additionally, the latent features derived from the convolutional encoder are quantized to perform contrastive tasks with the local contextual representations.
Quantization is the process of mapping continuous-valued feature vectors to a finite set of discrete-valued vectors (\textit{i.e.} codes) via a trainable codebook.
The codebook contains vectors where the latent features are replaced with the nearest code in the codebook, which is chosen via Gumbel-softmax layer~\citep{jang2016categorical}. 

Specifically, given a raw ECG input $X$, the convolutional encoder $f:X \Rightarrow Z$ produces the latent features $Z$.
Then, the Transformer $h:Z \Rightarrow C$ produces contextualized local representations $C$.
The quantization module $q: Z \rightarrow Q$ quantizes the latent features to $Q$.

During pre-training, the latent features at random time steps are quantized to $q$ as well as masked to $m$ before being fed into the Transformer (Shown by the part above the CNN layers in \figureref{fig:model}).
Then, the model is trained to maximize the cosine similarity between the local contextualized representation and its corresponding quantized feature at each masked time step.
This framework effectively captures salient local information from both local and quantized representations of each ECG signal.

\paragraph*{CMSC}
CMSC introduces a patient-specific method for effective self-supervised learning of unlabelled ECG data.
The temporal invariant nature of ECG recordings is exploited by defining adjacent temporal segments as positive pairs.
Specifically, given an $i$-th ECG recording with the duration of $S_i$ seconds, pairs of temporal segments are sampled by non-overlapping $S_i/2$ second segments (See the duration below the raw ECG signals in \figureref{fig:model}).
For simplicity, we fix the duration $S_i=10$ for each $i$-th ECG signal.
The adjacent segments are utilized as positive pairs, and other temporal segments from different ECG samples are used as negative pairs.
This contrastive learning method focuses on global representations of ECG recordings by comparing inter-relations of ECG recordings between patients.

\paragraph*{Wav2Vec 2.0+CMSC}
We integrate the Wav2Vec 2.0 model architecture and CMSC training scheme to learn both global and local contexts via contrastive learning.
The 12-channel raw ECG signals are segmented into $2 \ast N$ samples, where $N$ represents the original number of samples, and used as input for the local contrastive task shown in \figureref{fig:model}.
Similar to Wav2Vec 2.0, the model undergoes minimization of the local contrastive loss between a local contextualized representation and its quantized feature.
Specifically, given a local representation vector $\vec{c}_{t}$ at a masked time step $t$ and its quantized latent feature $\vec{q}_{t}$, the local contrastive loss is defined as
\begin{align}
    L_{t} &= -\log{
    \frac{e^{\textit{sim}(\vec{c}_t, \vec{q}_t)/\tau)}}{\sum_{\vec{q} \sim Q}{e^{\textit{sim}(\vec{c}_t, \vec{q})/\tau)}}}
    }\label{eq:local_loss}\\
    L_{local} &= \frac{1}{|M|} \sum_{t \in M}L_{t}
\end{align}
where \textit{sim}($\vec{a}, \vec{b}$) is the cosine similarity between two vectors $\vec{a}$ and $\vec{b}$, $Q$ is a set of quantized candidate latent features, which consist of $\vec{q}_t$ and quantized features from other masked time steps, and $M$ is a set of masked time steps.

Simultaneously, in order to learn global context together, temporal average pooling is applied to local contextual representations. (Shown by the upper part of Transformer block in \figureref{fig:model}).
Then, patient-specific noise contrastive estimation loss is utilized for learning the global representation of ECGs.
This global contrastive loss is defined as
\begin{align}
    L_{i,j} &= -\log{
    \frac{e^{\textit{sim}(\vec{g}_i, \vec{g}_j)/\tau)}}{\sum^{2N}_{k=1}{\mathbb{1}_{[k \neq i]} e^{\textit{sim}(\vec{g}_i, \vec{g}_k)/\tau)}}}
    }\\
    L_{global} &= \frac{1}{|P^+|} \sum_{i,j \in P^{+}}L_{i,j}\label{eq:global_loss}
\end{align}
where $\vec{g}_i=\frac{1}{T}\sum_{t \in T}\vec{c}_t$ stands for the global representation of the $i$-th sample, and $P^+$ denotes a set of indices of positive pairs in a batch.
Finally, the overall objective function of the integrated model can be described as Eq.~\ref{eq:loss}. The local contrastive learning task focuses on intra-relations within each ECG signal, while the global contrastive learning task exploits inter-relations between ECG recordings of different patients during training.
\begin{equation}
    L = L_{local} + L_{global}\label{eq:loss}
\end{equation}

\subsection{Random Lead Masking}
\label{subsec:rlm}
Although 12-lead ECG signals are the standard for various ECG-related downstream tasks, there is a growing demand for models that are able to provide robust performance for reduced-lead ECG signals \citep{green2007best}.
This is due to the limited accessibility of standard 12-lead ECG recordings, as attaching at least ten electrodes to the patient's body is not a trivial task.
In addition, the increasing popularity of ECG-measuring personal devices (\textit{i.e.} smart watches), which typically measure reduced-lead ECG signals, is likely to generate an even larger volume of reduced-lead ECGs.
Taking this into consideration, the obvious self-supervised learning strategy is to pre-train and fine-tune individual models for each set of leads.
However, this is practically infeasible since there are too many possible combinations with 12 leads.


Therefore, we propose Random Lead Masking (RLM) as an ECG-specific augmentation method, where we randomly mask each lead individually with the probability of $p=0.5$.
This strategy enables the 12-lead model to mimic the pre-training setting of using various lead combinations by stochastically masking random leads in every iteration.  
In other words, the model is exposed to diverse lead combinations at the pre-training stage, which reduces the dependence of representations on all 12 leads of ECG signals.
Thus, a single pre-trained model with RLM can show robust performance when being fine-tuned for downstream tasks with an arbitrary set of leads.

\section{Fine-tuning}
\label{sec:finetune}
After pre-training, we evaluate the models by fine-tuning on labeled data for two representative downstream tasks for ECG, namely cardiac arrhythmia classification and patient identification.
During the fine-tuning process, we apply temporal average pooling on the output of the Transformer (\textit{i.e.} contextualized local features $c_t$'s) to get a representation vector for each entire ECG signal.
Then, we add a randomly initialized fully connected layer after the temporal average pooling layer to perform the downstream tasks.

We experiment with five different combinations of available leads: 12-lead, 6-lead (I, II, III, aVF, aVL, aVR), 3-lead (I, II, V2), 2-lead (I, II), and 1-lead (I).
These lead combinations are identical to the setup used in \textit{PhysioNet/Computing in Cardiology Challenge 2021}~\citep{reyna2021cinc}, except for the 4-lead combination (I, II, III, V2).
It was excluded because of the correlation between the limb leads, where lead III can be obtained by the simple equation of lead I and II (\textit{i.e.} Lead III = $-$Lead I + Lead II).
Therefore, the 4-lead combination does not provide any additional information compared to the 3-lead combination.
We also added a 1-lead (I) experiment because many wearable devices such as smart watches measure only 1-lead ECG signals.


\paragraph*{Cardiac Arrhythmia Classification}
Cardiac arrhythmia classification is a task to predict the diagnosis of cardiac abnormalities of the heart. In this task, we classify ECG signals into 26 SNOMED-CT multi-labels which are scored according to~\cite{reyna2021cinc}.

For evaluation, we utilize the CinC Score, which is introduced in \emph{PhysioNet/Computing in Cardiology Challenge 2021}~\citep{reyna2021cinc}.
This metric is a weighted version of the traditional accuracy metric, where partial credit is given to misdiagnosis with similar symptoms as true diagnosis and penalty for misdiagnosis with very different symptoms to true diagnosis. 
This weighted score reflects the clinical reality that not all misdiagnoses are considered the same.
Models that output the correct classes for all the samples receive a score of 1, and models that always output the normal class (\textit{i.e.} no abnormality) receive a score of 0. The score ranges from -1 to 1.

\paragraph*{Patient Identification}
Patient identification is a task where the model is required to learn effective representations of the ECG signals so that the similarity between two different ECGs of the same patient are relatively high.

Here, we treat a fully connected layer as a collection of weight vectors for each patient (class).
That is, when we perform a classification task using the traditional softmax classifier, the feature vectors are trained to have a high similarity (\textit{i.e.} a high dot product) with its corresponding weight vector, and a low similarity (\textit{i.e.} a low dot product) with other weight vectors.
However, since the softmax classifier optimizes dot product similarity, it does not impel higher similarity among intra-patient ECGs and diversity among inter-patient ECGs.

To overcome this problem, we exploit the ArcFace~\citep{deng2019arcface} loss formulated as
\[
L = - \frac{1}{N} \sum_{i=1}^{N}\log\frac{e^{s \cos(\theta_{y_i} + m)}}{e^{s \cos(\theta_{y_i} + m)} + \sum_{j=1, j\neq y_i}^{N} e^{s \cos(\theta_{y_i})}}\\
\]
where $\theta_{y_i}$ is the angle between the feature vector and its corresponding weight vector, $m$ is the angular margin penalty to $\theta_{y_i}$, and $s$ is the scaling factor before applying softmax.
By optimizing cosine similarity with additional angular margin penalty, the ArcFace loss can maximize similarity for intra-patient ECGs while minimizing for inter-patient ECGs.

For evaluation, the test set is composed of two subsets which are called the gallery and probe sets.
The subsets consist of unique ECG samples but share the same patients.
For example, if the gallery set contains an ECG sample from patient $A$, the probe set must also contain ECG recording from the same patient $A$.
During testing, we discard the additional fully connected layer and only take the representation vectors of ECG samples in each subset into account.
Afterward, we calculate cosine similarities of all possible pairs between gallery and probe set, and define the closest pairs as having the same identity to each other.
The visualization of the training process of this task is provided in~\appendixref{apd:first}.
\renewcommand{\arraystretch}{1.25}

\begin{table*}[t]
\begin{adjustwidth}{-1in}{-1in}
\floatconts
  {tab:table1}
  {\caption{Test performances when pre-training a single 12-lead model and fine-tuning on the five lead combinations. In fine-tuning, we fill unavailable leads with zero, which is denoted as P-N-lead (Padded-N-lead). Here, W2V stands for Wav2Vec 2.0. We measure CinC Score for diagnosis classification (Dx.), and accuracy for patient identification (Id.). Mean and 95\% confidence interval are shown across 3 seeds. We highlight the best performances with \textbf{boldface} for each lead combination on two tasks.}}
  {\resizebox{\textwidth}{!}{
  \begin{tabular}{lcccccc}
  \toprule
    \multicolumn{2}{c}{\multirow{2}{*}{Methods}} &
    \multicolumn{5}{c}{Lead combinations}
    \\\cmidrule(lr){3-7}
      & & 12-lead & P-6-lead & P-3-lead & P-2-lead & P-1-lead
  \\\hline
  \midrule
  \textit{Baselines}
  \\\hline
  \multirow{2}{*}{Random Init.} & Dx. & $0.618 \pm 0.002$ & $0.522 \pm 0.010$ & $0.592 \pm 0.011$ & $0.525 \pm 0.020$ & $0.415 \pm 0.006$
  \\
  & Id. & $0.492 \pm 0.001$ & $0.346 \pm 0.004$ & $0.470 \pm 0.005$ & $0.358 \pm 0.003$ & $0.112 \pm 0.008$
  \\\hline
  \multirow{2}{*}{W2V} & Dx. & $0.714 \pm 0.011$ & $0.643 \pm 0.012$ & $0.676 \pm 0.011$ & $0.611 \pm 0.002$ & $0.525 \pm 0.016$
  \\
  & Id. & $0.492 \pm 0.004$ & $0.411 \pm 0.002$ & $0.470 \pm 0.001$ & $0.414 \pm 0.004$ & $0.247 \pm 0.003$
  \\\hline
  \multirow{2}{*}{CMSC} & Dx. & $0.625 \pm 0.006$ & $0.522 \pm 0.004$ & $0.575 \pm 0.010$ & $0.507 \pm 0.014$ & $0.406 \pm 0.006$
  \\
  & Id. & $0.513 \pm 0.006$ & $0.392 \pm 0.012$ & $0.510 \pm 0.003$ & $0.378 \pm 0.023$ & $0.227 \pm 0.008$
  \\\hline
  \multirow{2}{*}{3KG} & Dx. & $0.600 \pm 0.014$ & $0.515 \pm 0.005$ & $0.563 \pm 0.002$ & $ 0.505 \pm 0.014$ & $0.418 \pm 0.009$
  \\
  & Id. & $0.407 \pm 0.004$ & $0.320 \pm 0.004$ & $0.367 \pm 0.001$ & $0.310 \pm 0.003$ & $0.198 \pm 0.001$
  \\\hline
  \multirow{2}{*}{SimCLR(RLM)} & Dx. & $0.578 \pm 0.015$ & $0.497 \pm 0.002$ & $0.535 \pm 0.015$ & $0.484 \pm 0.004$ & $0.393 \pm 0.012$
  \\
  & Id. & $0.353 \pm 0.003$ & $0.289 \pm 0.004$ & $0.368 \pm 0.006$ & $0.304 \pm 0.003$ & $0.192 \pm 0.004$
  \\\hline
  \textit{Our methods}
  \\\hline
  \multirow{2}{*}{W2V+CMSC} & Dx. & $0.717 \pm 0.001$ & $0.616 \pm 0.012$ & $0.656 \pm 0.009$ & $0.586 \pm 0.010$ & $0.482 \pm 0.007$
  \\
  & Id. & $0.550 \pm 0.009$ & $0.437 \pm 0.002$ & $0.466 \pm 0.006$ & $0.410 \pm 0.003$ & $0.280 \pm 0.004$
  \\\hline
  \multirow{2}{*}{W2V+CMSC+RLM} & Dx. & $\bf 0.732 \pm 0.004$ & $\bf 0.662 \pm 0.011$ & $\bf 0.714 \pm 0.006$ & $\bf 0.656 \pm 0.010$ & $\bf 0.554 \pm 0.016$
  \\
  & Id. & $\bf 0.577 \pm 0.006$ & $\bf 0.459 \pm 0.007$ & $\bf 0.548 \pm 0.003$ & $\bf 0.457 \pm 0.005$ & $\bf 0.313 \pm 0.005$
  \\\hline
  \bottomrule
  \end{tabular}}}
  \end{adjustwidth}
\end{table*}

\renewcommand{\arraystretch}{1.25}

\begin{table*}[t]

\begin{adjustwidth}{-1in}{-1in}
\floatconts
  {tab:table2}
  {\caption{Test performances of various pre-training methods when pre-training and fine-tuning an individual model for each lead combination. We measure CinC Score for diagnosis classification (Dx.), and accuracy for patient identification (Id.). We highlight the best performances with \textbf{boldface}. Here, we bring 12-lead results from \tableref{tab:table1} as reference.}}
  {\resizebox{\textwidth}{!}{
  \begin{tabular}{lcccccc}
  \toprule
    \multicolumn{2}{c}{\multirow{2}{*}{Methods}} &
    \multicolumn{5}{c}{Lead combinations}
    \\\cmidrule(lr){3-7}
      & & 12-lead & 6-lead & 3-lead & 2-lead & 1-lead
  \\\hline
  \midrule
  \textit{Baselines}
  \\\hline
  \multirow{2}{*}{Random Init.} & Dx. & $0.618 \pm 0.002$ & $0.515 \pm 0.007$ & $0.596 \pm 0.013$ & $0.517 \pm 0.012$ & $0.406 \pm 0.009$
  \\
  & Id. & $0.492 \pm 0.001$ & $0.345 \pm 0.003$ & $0.470 \pm 0.004$ & $0.350 \pm 0.003$ & $0.114 \pm 0.005$
  \\\hline
  \multirow{2}{*}{W2V} & Dx. & $0.714 \pm 0.011$ & $\bf 0.696 \pm 0.007$ & $\bf 0.730 \pm 0.006$ & $\bf 0.710 \pm 0.011$ & $\bf 0.614 \pm 0.013$
  \\
  & Id. & $0.492 \pm 0.004$ & $0.478 \pm 0.002$ & $0.519 \pm 0.002$ & $0.498 \pm 0.004$ & $0.327 \pm 0.000$
  \\\hline
  \multirow{2}{*}{CMSC} & Dx. & $0.625 \pm 0.006$ & $0.491 \pm 0.009$ & $0.544 \pm 0.008$ & $0.497 \pm 0.018$ & $0.466 \pm 0.015$
  \\
  & Id. & $0.513 \pm 0.006$ & $0.396 \pm 0.004$ & $0.499 \pm 0.006$ & $0.421 \pm 0.005$ & $0.274 \pm 0.004$
  \\\hline
  \textit{Our methods}
  \\\hline
  \multirow{2}{*}{W2V+CMSC} & Dx. & $0.717 \pm 0.001$ & $0.689 \pm 0.017$ & $0.713 \pm 0.021$ & $0.696 \pm 0.013$ & $0.592 \pm 0.008$
  \\
  & Id. & $0.550 \pm 0.008$ & $\bf 0.514 \pm 0.002$ & $\bf 0.568 \pm 0.010$ & $\bf 0.537 \pm 0.003$ & $\bf 0.346 \pm 0.004$
  \\\hline
  \multirow{2}{*}{W2V+CMSC+RLM} & Dx. & $\bf 0.732 \pm 0.004$ & $0.666 \pm 0.011$ & $0.701 \pm 0.006$ & $0.660 \pm 0.003$ & -
  \\
  & Id. & $\bf 0.577 \pm 0.006$ & $0.495 \pm 0.006$ & $0.479 \pm 0.001$ & $0.447 \pm 0.009$ & -
  \\\hline
  \bottomrule
  \end{tabular}}}
  \end{adjustwidth}
\end{table*}

\renewcommand{\arraystretch}{1.25}

\begin{table*}[t]
\begin{adjustwidth}{-1in}{-1in}
\floatconts
  {tab:table3}
  {\caption{Test performances when applying random leads masking to W2V and CMSC. We measure CinC Score for diagnosis classification (Dx.), and accuracy for patient identification (Id.). If the performance is improved with RLM, we mark it with \textbf{boldface}.}}
  {\resizebox{\textwidth}{!}{
  \begin{tabular}{ccccccc}
  \toprule
    \multicolumn{2}{c}{\multirow{2}{*}{Methods}} &
    \multicolumn{5}{c}{Lead combinations}
    \\\cmidrule(lr){3-7}
      & & 12-lead & P-6-lead & P-3-lead & P-2-lead & P-1-lead
  \\\hline
  \midrule
  \multirow{2}{*}{W2V} & Dx. & $0.714 \pm 0.011$ & $0.643 \pm 0.012$ & $0.676 \pm 0.011$ & $0.611 \pm 0.002$ & $0.525 \pm 0.016$
  \\
  & Id. & $0.492 \pm 0.004$ & $0.411 \pm 0.002$ & $0.470 \pm 0.001$ & $0.414 \pm 0.004$ & $0.247 \pm 0.003$
  \\\cmidrule(lr){2-7}
  \multirow{2}{*}{W2V+RLM} & Dx. & $\bf 0.718 \pm 0.003$ & $\bf 0.649 \pm 0.007$ & $\bf 0.700 \pm 0.011$ & $\bf 0.655 \pm 0.009$ & $\bf 0.559 \pm 0.008$
  \\
  & Id. & $0.476 \pm 0.002$ & $\bf 0.421 \pm 0.002$ & $0.427 \pm 0.001$ & $0.398 \pm 0.004$ & $\bf 0.283 \pm 0.004$
  \\\hline
  \multirow{2}{*}{CMSC} & Dx. & $0.625 \pm 0.006$ & $0.522 \pm 0.004$ & $0.575 \pm 0.010$ & $0.507 \pm 0.014$ & $0.406 \pm 0.006$
  \\
  & Id. & $0.513 \pm 0.006$ & $0.392 \pm 0.012$ & $0.510 \pm 0.003$ & $0.378 \pm 0.023$ & $0.227 \pm 0.008$
  \\\cmidrule(lr){2-7}
  \multirow{2}{*}{CMSC+RLM} & Dx. & $\bf 0.697 \pm 0.002$ & $\bf 0.615 \pm 0.003$ & $\bf 0.665 \pm 0.006$ & $\bf 0.616 \pm 0.010$ & $\bf 0.499 \pm 0.006$
  \\
  & Id. & $\bf 0.565 \pm 0.002$ & $\bf 0.481 \pm 0.003$ & $\bf 0.554 \pm 0.006$ & $\bf 0.479 \pm 0.004$ & $\bf 0.309 \pm 0.003$
  \\\hline
  \bottomrule
  \end{tabular}}}
  \end{adjustwidth}
\end{table*}

\renewcommand{\arraystretch}{1.25}

\begin{table*}[t]

\begin{adjustwidth}{-1in}{-1in}
\floatconts
  {tab:table4}
  {\caption{Test performances when applying various augmentations to W2V+CMSC model. \textit{Physio}(4) stands for powerline noise, electromyographic noise noise, baseline wander, and baseline shift. \textit{Physio}(3) stands for powerline noise, electromyographic noise, and baseline wander. We measure CinC Score for diagnosis classification (Dx.), and accuracy for patient identification (Id.).}}
  {\resizebox{\textwidth}{!}{
  \begin{tabular}{lcccccc}
  \toprule
    \multirow{2}{*}{Augmentations} & &
    \multicolumn{5}{c}{Lead combinations}
    \\\cmidrule(lr){3-7}
      & & 12-lead & P-6-lead & P-3-lead & P-2-lead & P-1-lead
  \\\hline
  \midrule
  \multirow{2}{*}{RLM} & Dx. & $\bf 0.732 \pm 0.004$ & $\bf 0.662 \pm 0.011$ & $\bf 0.714 \pm 0.006$ & $\bf 0.656 \pm 0.010$ & $\bf 0.554 \pm 0.016$
  \\
  & Id. & $\bf 0.577 \pm 0.006$ & $0.459 \pm 0.007$ & $\bf 0.548 \pm 0.003$ & $0.457 \pm 0.005$ & $0.313 \pm 0.005$
  \\\hline
  \multirow{2}{*}{RLM+\textit{Physio}(4)} & Dx. & $0.697 \pm 0.003$ & $0.645 \pm 0.015$ & $0.674 \pm 0.004$ & $0.634 \pm 0.005$ & $0.562 \pm 0.001$
  \\
  & Id & $0.455 \pm 0.012$ & $0.409 \pm 0.004$ & $0.433 \pm 0.006$ & $0.377 \pm 0.002$ & $0.302 \pm 0.012$
  \\\hline
  \multirow{2}{*}{RLM+\textit{Physio}(3)} & Dx. & $0.706 \pm 0.007$ & $0.645 \pm 0.019$ & $0.691 \pm 0.002$ & $0.646 \pm 0.005$ & $0.540 \pm 0.000$
  \\
  & Id. & $0.558 \pm 0.008$ & $\bf 0.489 \pm 0.018$ & $0.528 \pm 0.007$ & $\bf 0.466 \pm 0.016$ & $\bf 0.322 \pm 0.005$
  \\\hline
  \multirow{2}{*}{\textit{Physio}(4)} & Dx. & $0.691 \pm0.002$ & $0.596 \pm 0.006$ & $0.644 \pm 0.006$ & $0.604 \pm 0.028$ & $0.496 \pm 0.001$
  \\
  & Id. & $0.450 \pm 0.009$ & $0.403 \pm 0.004$ & $0.450 \pm 0.005$ & $0.437 \pm 0.002$ & $0.271 \pm 0.002$
  \\\hline
  \multirow{2}{*}{\textit{Physio}(3)} & Dx. & $0.708 \pm 0.008$ & $0.610 \pm 0.002$ & $0.622 \pm 0.006$ & $0.554 \pm 0.005$ & $0.479 \pm 0.009$
  \\
  & Id. & $0.488 \pm 0.001$ & $0.401 \pm 0.008$ & $0.421 \pm 0.006$ & $0.373 \pm 0.006$ & $0.249 \pm 0.009$
  \\\hline
  \bottomrule
  \end{tabular}}}
  \end{adjustwidth}
\end{table*}

\section{Experiments}
\label{sec:exp}

In this section, we describe the datasets and implementation details of our work.
Then, we present our experimental results, where the experiments are as follows:


Firstly, we pre-train our proposed models and other baselines on the massive ECG dataset using all 12 leads. Then, for two downstream tasks, we fine-tune the single 12-lead model on the five reduced-lead combinations by filling unavailable leads with zero.

Secondly, we pre-train and fine-tune our proposed models and other baselines for each lead combination, respectively.
We conduct this experiment to show the theoretical upper-bound performance of our first experiment.
This experiment requires a massive amount of training hours and computational resources since we individually pre-train the models for each set of leads.


Lastly, we conduct an ablation study regarding RLM. The objective of this study is to observe the outcome of utilizing RLM with other baseline models.
The enhanced performance using RLM shows the effectiveness of RLM as an ECG-specific augmentation for self-supervised contrastive learning.

Note that we do not perform the second and third experiments with 3KG since the primary objective of 3KG lies in converting ECGs into VCGs and applying stochastic 3-D perturbations to them to generate positive pairs.
Using fewer ECG leads during conversion to VCG will fail to construct the 3-D structure, and the perturbations will not have any contextual meaning.

\subsection{Datasets}
\label{sec:dataset}
\paragraph*{PhysioNet 2021}
We conduct pre-training experiments on six datasets in \textbf{PhysioNet 2021} \citep{reyna2021cinc} which are CPSC, CPSC-Extra, PTB-XL, Georgia, Ningbo, and Chapman.
Each sample has a sampling frequency of 500 Hz and ranges between 5 and 144 seconds.
As mentioned in \sectionref{sec:contrastive}, we repeatedly crop $S_i=10$ second temporal segments for each $i$-th data sample.
Additionally, we split each 10-second sample into two $S_i/2=5$ second segments for processing data easily for the global contrastive task.
This leads to 189,051 samples of 12-lead ECG recordings, each of which has a sample size of 2500. 

In fine-tuning for cardiac arrhythmia classification, we utilize the CPSC and Georgia datasets since these two datasets are used for evaluation in \emph{PhysioNet/Computing in Cardiology Challenge 2021}~\citep{reyna2021cinc}.\footnote{To be more specific, the data used for evaluation in CinC Challenge 2021 has not been shared publicly. However, since the hidden evaluation datasets are partly composed of these two datasets (CPSC and Georgia), we used them as our fine-tuning dataset for cardiac arrhythmia classification.}
We split the samples into training, validation, and test sets according to an 8:1:1 ratio, yielding 32640, 4079, and 4079 samples, respectively.
The validation and test sets are excluded from the pre-training dataset.

In fine-tuning for the patient identification task, we use the pre-training dataset for training and validation.
The train and validation sets are composed of 147,444 and 17,670 samples, which are 8:2 ratios of the pre-training dataset.
Note that most PhysioNet 2021 datasets do not contain patient identity information, so we consider the segmented samples from the same ECG as having the same identity.
After the fine-tuning, however, we test the model performance on PTB-XL, which does contain patient identity, as we will describe below.

\paragraph*{PTB-XL}
To appropriately test the model for the patient identification task, we utilize \textbf{PTB-XL} dataset~\citep{wagner2020physionet}, which is a subset of PhysioNet 2021 but has a patient id for each ECG sample.
Thus, we can identify ECG samples with different sessions, which is a more appropriate way to evaluate the model.
We select patients that have at least 2 ECG sessions and randomly crop them into 5 seconds.
Then, we randomly choose two ECG sessions for each unique patient, and use them as the gallery and probe sets, respectively. As a result, we retain 2127 unique patient pairs of 12-lead ECG recordings, which are excluded from the pre-training dataset.

\subsection{Implementation Details}
We implemented our experiments with FAIRSEQ framework~\citep{ott2019fairseq}, which is based on PyTorch~\citep{NEURIPS2019_9015}.
The feature extractor is composed of 4 blocks, each of which consists of a convolutional layer followed by a layer normalization~\citep{ba2016layer}, and a GELU activation function~\citep{hendrycks2016gaussian}. The convolutional layers in each block have 256 channels with strides of 2 and kernel lengths of 2. Furthermore, the Transformer setup was consistent with the BERT-BASE model~\citep{kenton2019bert}, where the number of transformer block layers is 12, the dimension of the model is 768, the number of self-attention heads is 12, and the dimension of the feed-forward network is 3,072.
When fine-tuning, the additional fully connected layer projects the representation vectors to the number of classes of each task.
For the identification task, the class stands for the unique patients in the training dataset.

For the configurations in local contrastive learning, we follow hyper-parameter settings of the original \textbf{Wav2Vec 2.0}.
We select each token as the start of the span to be masked with the probability of 0.065 and mask the subsequent 10 time-steps.
In addition, the quantization module contains two groups of 320 codes.
For \textbf{ArcFace} in identification, we use the scaling factor of $s=192$, and margin of $m=1.0$.

We optimize the model using Adam~\citep{kingma2014adam} with the learning rate of $5 \times 10^{-5}$ for pre-training and fine-tuning on the classification task, and $3 \times 10^{-5}$ for fine-tuning on identification task.
For pre-training, the batch size is 10-second 512 ECG samples, which are segmented into 5-second 1024 ECG samples, on 4 \textbf{RTX A6000} GPUs, giving a training time of 24 hours.
For fine-tuning, the batch size is 5-second 128 ECG samples on a single \textbf{RTX 3090} GPU, resulting in a training time of 4 hours for cardiac arrhythmia classification and 18 hours for patient identification, respectively.

\subsection{Fine-tuning a Single 12-lead Model on Padded Leads}
\label{subsec:res1}
In \tableref{tab:table1}, we represent the experimental results when pre-training a single 12-lead model and fine-tuning on various lead combinations. In other words, a single pre-trained model by 12-lead is fine-tuned on reduced-leads by filling unavailable leads with zero. 

The experimental results show that our proposed method, W2V+CMSC+RLM, outperforms other state-of-the-art methods that learn local or global context disjointly.
For the classification task, W2V+CMSC+RLM shows an average of $0.0298$ and $0.1366$ improvement in CinC Score compared to W2V and CMSC, respectively, across all lead combinations.
Similarly, for the identification task, W2V+CMSC+RLM achieves an average of $6.4\%p$ and $6.67\%p$ increase in accuracy compared to W2V and CMSC, respectively, across all lead combinations.

Additionally, we can also see that the performances on P-3-lead are consistently higher than P-2-lead or P-6-lead for all methods.
This is because any two limb leads can be used to calculate the other four leads since they are measured on the same frontal (coronal) plane.
Therefore, 2-lead and 6-lead contain the same amount of ECG information.
On the other hand, 3-lead includes a precordial lead (V2) together with two limb leads (I, II), so that 3-lead has additional ECG information compared to 2-lead or 6-lead.

In addition, comparing the results of W2V+CMSC+RLM with W2V+CMSC, there exists a significant gap for all lead combinations. This shows the benefit of utilizing RLM with W2V+CMSC when pre-training for 12-lead ECG data.

\subsection{Pre-training and Fine-tuning for Each Lead Combination}
\label{subsec:res2}

The fine-tuning results of the individually pre-trained models for each lead combination are shown in~\tableref{tab:table2}.
For the classification task, W2V shows the best score for all lead combinations other than 12-lead.
As mentioned in \sectionref{sec:finetune}, this classification task focuses on local contextual representations for accurate diagnosis.
Therefore, the local contrastive method, W2V, is able to show stronger performance compared to global and local contrastive method, W2V+CMSC+RLM.

For the identification task, W2V+CMSC shows optimal performance for all other lead combinations other than 12-lead.
In this case, global and local representations learned for each lead combination were able to outperform either local or global representations.

In this experimental setup, the performance of W2V+CMSC+RLM was consistently worse than W2V+CMSC for all reduced-lead combinations (6-lead, 3-lead, 2-lead).
We hypothesize that this is due to using fewer leads at pre-training with RLM, which harms the expressiveness of the model compared to when using all 12 leads.
In other words, since we reduced available leads, the possible combinations of leads significantly decreased together.
Since the efficacy of RLM originates from that the model can learn diverse combinations of leads at the pre-training stage, reducing available leads is detrimental to the model from learning robust representations for an arbitrary set of leads.



\subsection{Ablation of RLM}

As shown in \tableref{tab:table3}, W2V+RLM shows improved performance for all lead combinations on the classification task.
However, the performance decreases for some lead combinations on the identification task when applying RLM.
We speculate that the main reason for this outcome is because using RLM with W2V enhances the W2V's capability to catch local context rather than global context.
Accordingly, compared to W2V, W2V+RLM shows superior performance in classification task that requires local context but decreased performance in identification task that requires global context.

On the other hand, CMSC+RLM shows drastically increased performance compared to CMSC for all cases.
We hypothesize that the contrastive task in the original CMSC, which is to maximize agreement between temporal adjacent ECG segments, is too simple for the model to learn valuable representations.
As mentioned in \sectionref{subsec:rlm}, RLM helps the model learn diverse combinations of leads, which means the model can explore massive amounts of augmented ECG samples for each patient during pre-training. 
This ensures CMSC learns useful representations of patients showing much better performance for all lead combinations on two downstream tasks.

\section{Discussion and Future Work}
\label{sec:discussion}
In this work, we presented a self-supervised learning method for global and local ECG representations by combining existing contrastive learning schemes, W2V and CMSC.
In addition, we proposed RLM, an augmentation technique that masks each lead randomly at the pre-training stage, in order to obtain a single pre-trained model robust against when all 12-leads are not available at the fine-tuning stage.
The experimental results showed that our proposed method, W2V+CMSC+RLM, outperforms the state-of-the-art methods.

\paragraph{Augmentations}
We introduced random lead masking as an augmentation method to allow a single pre-trained model to be fine-tuned successfully with an arbitrary number of lead combinations.
However, there are many previous and ongoing research on ECG-specific augmentation methods that exploit relations between leads that can potentially improve the performance of our proposed method.
In \cite{mehari2021self}, four different ECG-specific augmentation methods are introduced, which are baseline wander, powerline noise, electromyographic noise, and baseline shift.
Details of these augmentations are provided in \appendixref{apd:second}.

We experimented W2V+CMSC with applying these four augmentation methods and RLM (RLM+\textit{Physio}(4)).
As shown in \tableref{tab:table4}, it shows a decrease in performance compared to RLM, across all lead combinations except for P-1-lead on Dx.
We speculate that the reason is because our dataset does not contain a high level of noise, but augmentation method such as baseline shift produces too many alterations to the original ECG samples for the model to train properly.
Accordingly, we conducted experiments excluding baseline shift, and results are stated as RLM+\textit{Physio}(3).
Although the performance improved without the baseline shift, using only RLM as an augmentation still shows better performance in most cases.
Furthermore, when applying only the ECG-specific augmentations without RLM (\textit{i.e.} \textit{Physio}(4) and \textit{Physio}(3)), the performance mostly decreased, especially showing a significant gap on reduced leads.
We leave the exploration of more advanced ECG-specific augmentations as our future work.

\paragraph{Limitations}
Our experiments contain some limitations.
First, although all baselines and our model used the same set of hyperparameters for fair comparison, we did not investigate optimal hyperparameters in-depth, such as the probability of RLM (we used 0.5 in all experiments), leaving some room for squeezing out maximum possible performance. 
Second, even though our dataset contains a large amount of ECG samples that are collected from multiple healthcare facilities, all samples we used have a sampling rate of 500 Hz.
Given that there are many ECGs recorded with various sampling rates, we plan to expand our work to frequency-agnostic self-supervised learning of ECGs.




\section*{Institutional Review Board (IRB)}
This research does not require IRB approval.

\acks{
This work was supported by Institute of Information \& Communications Technology Planning \& Evaluation (IITP) grant (No.2019-0-00075, Artificial Intelligence Graduate School Program(KAIST)) and National Research Foundation of Korea (NRF) grant (NRF-2020H1D3A2A03100945) funded by the Korea government (MSIT) and by Medical AI Inc.
}

\clearpage
\bibliography{jmlr-sample}

\begin{thebibliography}{24}
\providecommand{\natexlab}[1]{#1}
\providecommand{\url}[1]{\texttt{#1}}
\expandafter\ifx\csname urlstyle\endcsname\relax
  \providecommand{\doi}[1]{doi: #1}\else
  \providecommand{\doi}{doi: \begingroup \urlstyle{rm}\Url}\fi

\bibitem[Ba et~al.(2016)Ba, Kiros, and Hinton]{ba2016layer}
Jimmy~Lei Ba, Jamie~Ryan Kiros, and Geoffrey~E Hinton.
\newblock Layer normalization.
\newblock \emph{arXiv preprint arXiv:1607.06450}, 2016.

\bibitem[Baevski et~al.(2020)Baevski, Zhou, Mohamed, and
  Auli]{baevski2020wav2vec}
Alexei Baevski, Henry Zhou, Abdelrahman Mohamed, and Michael Auli.
\newblock wav2vec 2.0: A framework for self-supervised learning of speech
  representations.
\newblock \emph{arXiv preprint arXiv:2006.11477}, 2020.

\bibitem[Chen et~al.(2020)Chen, Kornblith, Norouzi, and Hinton]{chen2020simple}
Ting Chen, Simon Kornblith, Mohammad Norouzi, and Geoffrey Hinton.
\newblock A simple framework for contrastive learning of visual
  representations.
\newblock In \emph{International conference on machine learning}, pages
  1597--1607. PMLR, 2020.

\bibitem[Deng et~al.(2019)Deng, Guo, Xue, and Zafeiriou]{deng2019arcface}
Jiankang Deng, Jia Guo, Niannan Xue, and Stefanos Zafeiriou.
\newblock Arcface: Additive angular margin loss for deep face recognition.
\newblock In \emph{Proceedings of the IEEE/CVF Conference on Computer Vision
  and Pattern Recognition}, pages 4690--4699, 2019.

\bibitem[Gopal et~al.(2021)Gopal, Han, Raghupathi, Ng, Tison, and
  Rajpurkar]{gopal20213kg}
Bryan Gopal, Ryan~W Han, Gautham Raghupathi, Andrew~Y Ng, Geoffrey~H Tison, and
  Pranav Rajpurkar.
\newblock 3kg: Contrastive learning of 12-lead electrocardiograms using
  physiologically-inspired augmentations.
\newblock \emph{arXiv preprint arXiv:2106.04452}, 2021.

\bibitem[Green et~al.(2007)Green, Ohlsson, Forberg, Bj{\"o}rk, Edenbrandt, and
  Ekelund]{green2007best}
Michael Green, Mattias Ohlsson, Jakob~Lundager Forberg, Jonas Bj{\"o}rk, Lars
  Edenbrandt, and Ulf Ekelund.
\newblock Best leads in the standard electrocardiogram for the emergency
  detection of acute coronary syndrome.
\newblock \emph{Journal of electrocardiology}, 40\penalty0 (3):\penalty0
  251--256, 2007.

\bibitem[Grill et~al.(2020)Grill, Strub, Altch{\'e}, Tallec, Richemond,
  Buchatskaya, Doersch, Pires, Guo, Azar, et~al.]{grill2020bootstrap}
Jean-Bastien Grill, Florian Strub, Florent Altch{\'e}, Corentin Tallec,
  Pierre~H Richemond, Elena Buchatskaya, Carl Doersch, Bernardo~Avila Pires,
  Zhaohan~Daniel Guo, Mohammad~Gheshlaghi Azar, et~al.
\newblock Bootstrap your own latent: A new approach to self-supervised
  learning.
\newblock \emph{arXiv preprint arXiv:2006.07733}, 2020.

\bibitem[Hendrycks and Gimpel(2016)]{hendrycks2016gaussian}
Dan Hendrycks and Kevin Gimpel.
\newblock Gaussian error linear units (gelus).
\newblock \emph{arXiv preprint arXiv:1606.08415}, 2016.

\bibitem[Jacob et~al.(2019)Jacob, Ming-Wei, Kenton, and
  Kristina]{kenton2019bert}
Devlin Jacob, Chang Ming-Wei, Lee Kenton, and Toutanova Kristina.
\newblock Bert: Pre-training of deep bidirectional transformers for language
  understanding.
\newblock In \emph{Proceedings of NAACL-HLT}, pages 4171--4186, 2019.

\bibitem[Jang et~al.(2016)Jang, Gu, and Poole]{jang2016categorical}
Eric Jang, Shixiang Gu, and Ben Poole.
\newblock Categorical reparameterization with gumbel-softmax.
\newblock \emph{arXiv preprint arXiv:1611.01144}, 2016.

\bibitem[Kachuee et~al.(2018)Kachuee, Fazeli, and Sarrafzadeh]{kachuee2018ecg}
Mohammad Kachuee, Shayan Fazeli, and Majid Sarrafzadeh.
\newblock Ecg heartbeat classification: A deep transferable representation.
\newblock In \emph{2018 IEEE International Conference on Healthcare Informatics
  (ICHI)}, pages 443--444. IEEE, 2018.

\bibitem[Kingma and Ba(2014)]{kingma2014adam}
Diederik~P Kingma and Jimmy Ba.
\newblock Adam: A method for stochastic optimization.
\newblock \emph{arXiv preprint arXiv:1412.6980}, 2014.

\bibitem[Kiyasseh et~al.(2021)Kiyasseh, Zhu, and Clifton]{kiyasseh2021clocs}
Dani Kiyasseh, Tingting Zhu, and David~A Clifton.
\newblock Clocs: Contrastive learning of cardiac signals across space, time,
  and patients.
\newblock In \emph{International Conference on Machine Learning}, pages
  5606--5615. PMLR, 2021.

\bibitem[Labati et~al.(2019)Labati, Mu{\~n}oz, Piuri, Sassi, and
  Scotti]{labati2019deep}
Ruggero~Donida Labati, Enrique Mu{\~n}oz, Vincenzo Piuri, Roberto Sassi, and
  Fabio Scotti.
\newblock Deep-ecg: Convolutional neural networks for ecg biometric
  recognition.
\newblock \emph{Pattern Recognition Letters}, 126:\penalty0 78--85, 2019.

\bibitem[Li et~al.(2020)Li, Pang, Wang, and Li]{li2020toward}
Yazhao Li, Yanwei Pang, Kongqiao Wang, and Xuelong Li.
\newblock Toward improving ecg biometric identification using cascaded
  convolutional neural networks.
\newblock \emph{Neurocomputing}, 391:\penalty0 83--95, 2020.

\bibitem[Mehari and Strodthoff(2021)]{mehari2021self}
Temesgen Mehari and Nils Strodthoff.
\newblock Self-supervised representation learning from 12-lead ecg data.
\newblock \emph{arXiv preprint arXiv:2103.12676}, 2021.

\bibitem[Oord et~al.(2018)Oord, Li, and Vinyals]{oord2018representation}
Aaron van~den Oord, Yazhe Li, and Oriol Vinyals.
\newblock Representation learning with contrastive predictive coding.
\newblock \emph{arXiv preprint arXiv:1807.03748}, 2018.

\bibitem[Ott et~al.(2019)Ott, Edunov, Baevski, Fan, Gross, Ng, Grangier, and
  Auli]{ott2019fairseq}
Myle Ott, Sergey Edunov, Alexei Baevski, Angela Fan, Sam Gross, Nathan Ng,
  David Grangier, and Michael Auli.
\newblock fairseq: A fast, extensible toolkit for sequence modeling.
\newblock In \emph{Proceedings of NAACL-HLT 2019: Demonstrations}, 2019.

\bibitem[Paszke et~al.(2019)Paszke, Gross, Massa, Lerer, Bradbury, Chanan,
  Killeen, Lin, Gimelshein, Antiga, Desmaison, Kopf, Yang, DeVito, Raison,
  Tejani, Chilamkurthy, Steiner, Fang, Bai, and Chintala]{NEURIPS2019_9015}
Adam Paszke, Sam Gross, Francisco Massa, Adam Lerer, James Bradbury, Gregory
  Chanan, Trevor Killeen, Zeming Lin, Natalia Gimelshein, Luca Antiga, Alban
  Desmaison, Andreas Kopf, Edward Yang, Zachary DeVito, Martin Raison, Alykhan
  Tejani, Sasank Chilamkurthy, Benoit Steiner, Lu~Fang, Junjie Bai, and Soumith
  Chintala.
\newblock Pytorch: An imperative style, high-performance deep learning library.
\newblock In \emph{Advances in Neural Information Processing Systems 32}, pages
  8024--8035. Curran Associates, Inc., 2019.

\bibitem[Peters et~al.(2018)Peters, Neumann, Iyyer, Gardner, Clark, Lee, and
  Zettlemoyer]{peters1802deep}
ME~Peters, M~Neumann, M~Iyyer, M~Gardner, C~Clark, K~Lee, and L~Zettlemoyer.
\newblock Deep contextualized word representations. arxiv 2018.
\newblock \emph{arXiv preprint arXiv:1802.05365}, 12, 2018.

\bibitem[Reyna et~al.(2021)Reyna, Sadr, Gu, Perez~Alday, Liu, Seyedi, Shah, and
  Clifford]{reyna2021cinc}
Matthew Reyna, Nadi Sadr, Annie Gu, Erick~Andres Perez~Alday, Chengyu Liu,
  Salman Seyedi, Amit Shah, and Gari Clifford.
\newblock Will two do? varying dimensions in electrocardiography: the
  {PhysioNet/Computing in Cardiology Challenge 2021}.
\newblock \emph{Computing in Cardiology 2021}, 48:\penalty0 1--4, 2021.

\bibitem[Vaswani et~al.(2017)Vaswani, Shazeer, Parmar, Uszkoreit, Jones, Gomez,
  Kaiser, and Polosukhin]{vaswani2017attention}
Ashish Vaswani, Noam Shazeer, Niki Parmar, Jakob Uszkoreit, Llion Jones,
  Aidan~N Gomez, {\L}ukasz Kaiser, and Illia Polosukhin.
\newblock Attention is all you need.
\newblock In \emph{Advances in neural information processing systems}, pages
  5998--6008, 2017.

\bibitem[Wagner et~al.(2020)Wagner, Strodthoff, Bousseljot, Samek, and
  Schaeffter]{wagner2020physionet}
Patrick Wagner, Nils Strodthoff, Ralf-Dieter Bousseljot, Wojciech Samek, and
  Tobias Schaeffter.
\newblock {PTB-XL}, a large publicly available electrocardiography dataset
  (version 1.0.1.
\newblock \emph{PhysioNet}, 2020.
\newblock \doi{https://doi.org/10.13026/x4td-x982}.

\bibitem[Yan et~al.(2019)Yan, Liang, Zhang, and Liu]{yan2019fusing}
Genshen Yan, Shen Liang, Yanchun Zhang, and Fan Liu.
\newblock Fusing transformer model with temporal features for ecg heartbeat
  classification.
\newblock In \emph{2019 IEEE International Conference on Bioinformatics and
  Biomedicine (BIBM)}, pages 898--905. IEEE, 2019.

\end{thebibliography}

\clearpage
\appendix
\onecolumn
\renewcommand{\thefigure}{\arabic{figure}A}

\section{Visualization of Training Process for Patient 
Identification}
\label{apd:first}

\begin{figure}[htbp]
\floatconts
    {fig:arcface}
    {\caption{Training process of patient identification task utilizing ArcFace loss. When validating or testing, we discard weight vectors and only get the normalized global representation vectors of ECG samples. Then, for each probe sample, we calculate cosine similarities with all the gallery samples, where the closest pair is considered to share the same identity.}}
    {\includegraphics[width=1.0\linewidth]{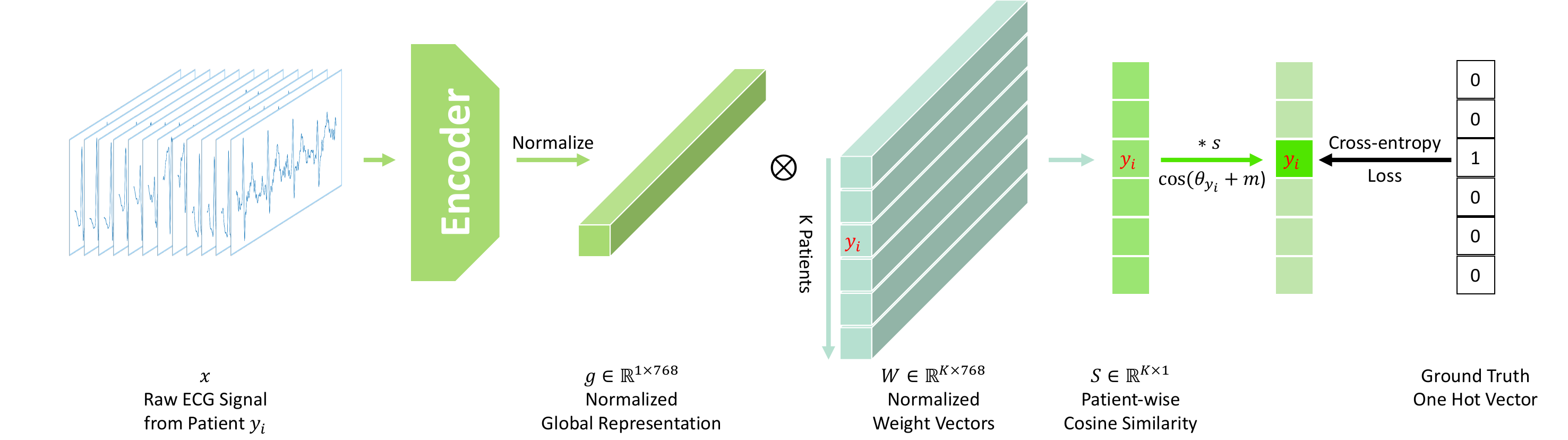}}
\end{figure}

\clearpage
\renewcommand{\thefigure}{\arabic{figure}B}

\section{Visualization of ECG-specific Augmentations}
\label{apd:second}
In this section, we describe the ECG-specific physiological augmentations applied in our work, which are initially introduced in ~\cite{mehari2021self}.
We also provide visualizations of these perturbations in \figureref{fig:augs}.

\begin{enumerate}
    \item \textbf{Powerline Noise}: Powerline interference is added to original signals. The noise is defined as
        \[
        n(t) = \sum^K_{k=1}a_k\cos(2\pi tkf_n+\phi)
        \]
        where $K=1$, $f_n=50$ Hz, $a_k \sim \mathcal{U}(0, 0.5)$, $\phi \sim \mathcal{U}(0, 2\pi)$.
    
    \item \textbf{Electromyographic Noise}: Electromyographic is a high-frequency noise commonly from muscle contraction. The noise is defined as
    \[
    n(t) \sim \mathcal{N}(0, a_t^2)
    \]
    where $a_t \sim \mathcal{U}(0, 0.5)$.
    
    \item \textbf{Baseline Wander}: Baseline wandering is considered as a low frequency artifact of ECG. The noise is defined as
    \[
    n(t) = C\sum^K_{k=1}a_k\cos(2\pi tk\Delta f + \phi_k)
    \]
    where $K=3$, $C \sim \mathcal{N}(1, 0.5^2)$, $a_k \sim \mathcal{U}(0, 0.5)$, $\Delta f \sim \mathcal{U}(0.01, 0.2)$, and $\phi_k \sim \mathcal{U}(0, 2\pi)$.
    
    \item \textbf{Baseline Shift}: Baseline shift changes baseline of random sampled steps from original signals. The noise is defined as
    \[
    n \sim \mathcal{U}(-0.5, 0.5)
    \]
    \[
    n(t) =
    \begin{cases}
        n & \text{if $t_{start} \leq t \leq t_{end}$} \\
        0 & \text{otherwise}
    \end{cases}
    \]
    where $[t_{start}, t_{end}]$ is a randomly sampled span of original signals.
\end{enumerate}

\begin{figure}[htbp]
\floatconts
  {fig:augs}
  {\caption{Visualizations of ECG-specific augmentations.}}
  {%
    \subfigure[Original]{\label{fig:original}%
      \includegraphics[width=0.45\linewidth]{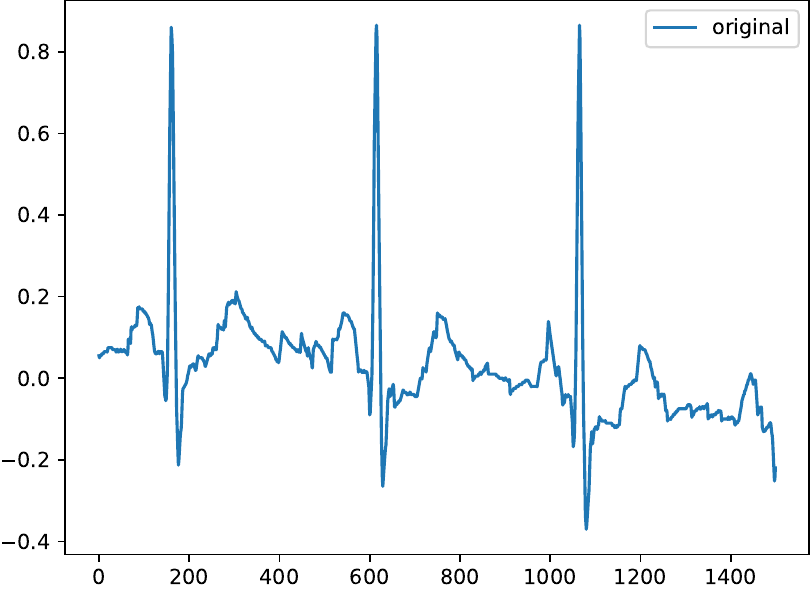}}%
    \\
    \subfigure[Powerline noise]{\label{fig:pln}%
      \includegraphics[width=0.45\linewidth]{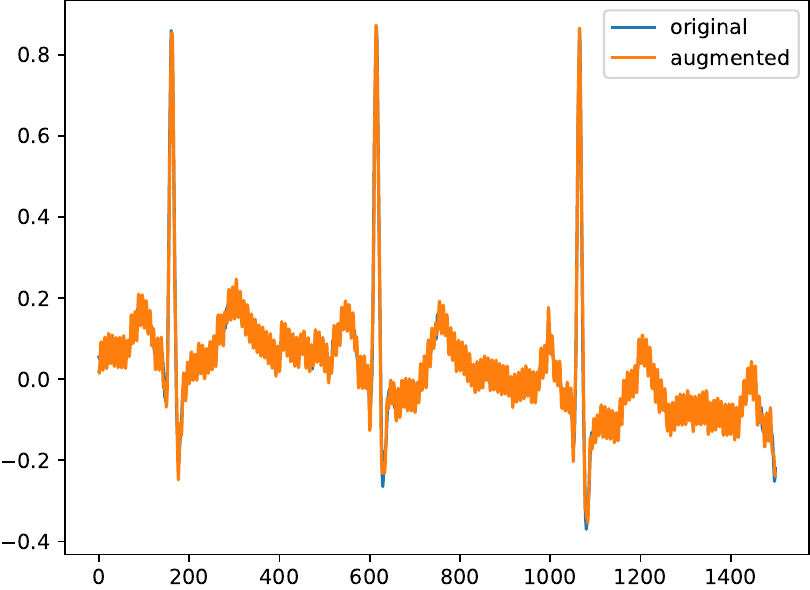}}
    \qquad
    \subfigure[Electromyographic noise]{\label{fig:emg}%
      \includegraphics[width=0.45\linewidth]{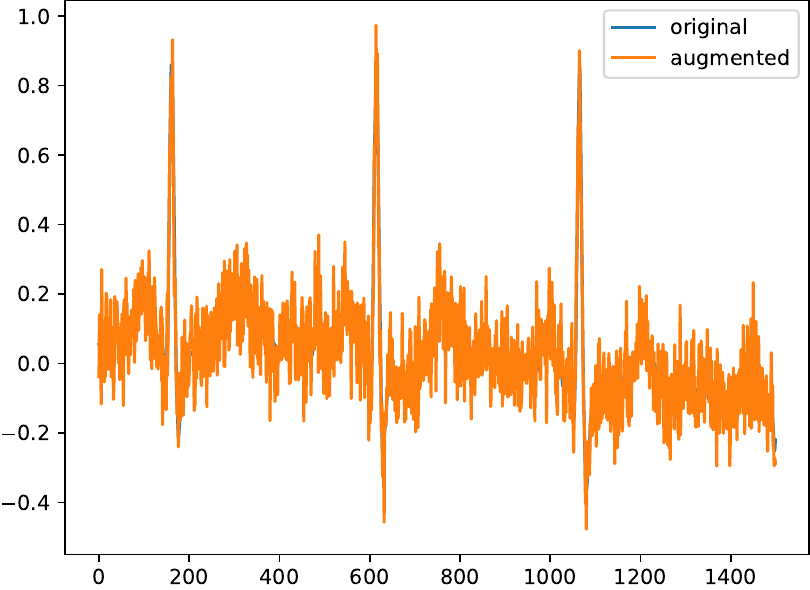}}
    \qquad
    \subfigure[Baseline wander]{\label{fig:bw}%
      \includegraphics[width=0.45\linewidth]{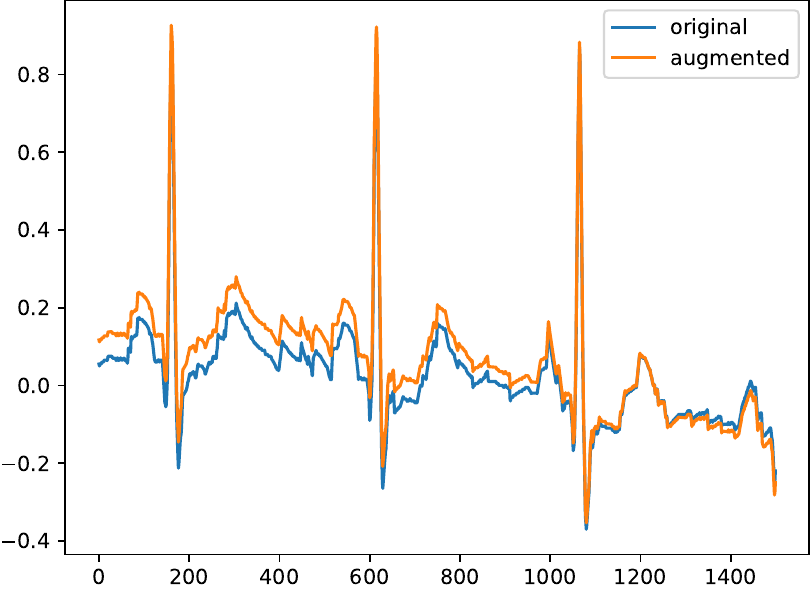}}
    \qquad
    \subfigure[Baseline shift]{\label{fig:bs}%
      \includegraphics[width=0.45\linewidth]{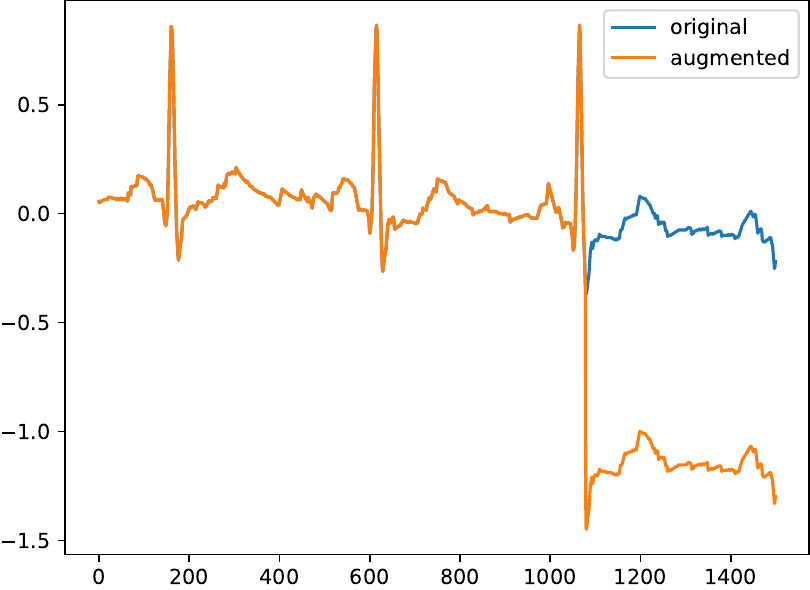}}

  }
\end{figure}




\end{document}